\documentclass[twoside]{article}

\usepackage[accepted]{aistats2024}
\usepackage{hyperref}
\usepackage{url}

\usepackage{booktabs}       
\usepackage{amsfonts}       

\usepackage{amsmath,amsthm}
\usepackage{graphicx}       
\usepackage{mathtools}
\usepackage{csquotes}
\usepackage{xcolor}
\usepackage{bbm}
\usepackage{bm}
\usepackage{subcaption}
\usepackage{enumitem,nicefrac}

\DeclareMathOperator*{\prob}{\mathbb{P}}
\DeclareMathOperator*{\lab}{\mathcal{Y}}

\DeclareMathOperator*{\plab}{\prob(\mathcal{Y})}
\DeclareMathOperator*{\pplab}{\prob(\prob(\mathcal{Y}))}
\DeclareMathOperator*{\Dir}{\text{Dir}}
\DeclareMathOperator*{\argmax}{\text{argmax}}

\newtheorem{theorem}{Theorem}[section]
\newtheorem{lemma}{Lemma}[section]
\newtheorem{prop}{Proposition}[section]
\newtheorem{corollary}{Corollary}[section]

\newtheorem{defi}{Definition}[section]

\newcommand{\TU}{\operatorname{TU}}
\newcommand{\EU}{\operatorname{EU}}
\newcommand{\AU}{\operatorname{AU}}

\allowdisplaybreaks

\newenvironment{sizeddisplay}[1]
 {\par\nopagebreak#1\noindent\ignorespaces}
 {\nopagebreak\ignorespacesafterend}

\runningtitle{Second-Order Uncertainty Quantification}
\runningauthor{Sale, Bengs, Caprio and Hüllermeier}

\usepackage[round]{natbib}

\begin{document}
	%
	
	%
	
	\twocolumn[
	
	\aistatstitle{Second-Order Uncertainty Quantification: \\ A Distance-Based Approach}
	
	\aistatsauthor{\href{mailto:<yusuf.sale@ifi.lmu.de>?Subject=Your Paper}{ Yusuf Sale }\textsuperscript{\textnormal{1,2}} \And Viktor Bengs\textsuperscript{\textnormal{1,2}} \And  Michele Caprio\textsuperscript{\textnormal{3}} \And Eyke Hüllermeier\textsuperscript{\textnormal{1,2}}}
	\vspace{0.2cm}
	\aistatsaddress{\textsuperscript{\textnormal{1}}Institute of Informatics, LMU Munich \\
    \textsuperscript{\textnormal{2}}Munich Center for Machine Learning (MCML) \\
    \textsuperscript{\textnormal{3}}PRECISE Center, University of Pennsylvania} 
    ]
	
	\begin{abstract}
		In the past couple of years, various approaches to representing and quantifying different types of predictive uncertainty in machine learning, notably in the setting of classification, have been proposed on the basis of second-order probability distributions, i.e., predictions in the form of distributions on probability distributions. A completely conclusive solution has not yet been found, however, as shown by recent criticisms of commonly used uncertainty measures associated with second-order distributions, identifying undesirable theoretical properties of these measures. In light of these criticisms, we propose a set of formal criteria that meaningful uncertainty measures for predictive uncertainty based on second-order distributions should obey. Moreover, we provide a general framework for developing uncertainty measures to account for these criteria, and offer an instantiation based on the Wasserstein distance, for which we prove that all criteria are satisfied. 
	\end{abstract}
	
	\section{INTRODUCTION}
	
	The need for representing and quantifying uncertainty in machine learning (ML) -- particularly in supervised learning scenarios -- has become more and more obvious in the recent past \citep{hullermeier2021aleatoric}.
	This is largely due to the increasing use of AI-driven systems in safety-critical real-world applications having stringent safety requirements, such as healthcare \citep{lambrou2010reliable, senge_2014_ReliableClassificationLearning, yang2009using} and socio-technical systems \citep{varshney2016engineering,varshney2017safety}.
    Dealing appropriately with uncertainty is a fundamental necessity in all these domains. 
    
	Broadly, uncertainties are categorized as \textit{aleatoric}, stemming from inherent data variability, and \textit{epistemic}, which arises from a model's incomplete knowledge of the data-generating process. By its very nature, epistemic uncertainty (EU) -- often being characterized as \textit{reducible} -- can be decreased with further information. In contrast, aleatoric uncertainty (AU), rooted in the data generating process itself, is fixed and cannot be mitigated \citep{hullermeier2021aleatoric}. The distinction between these uncertainty types has been a subject of keen interest in recent ML and statistical research \cite{gruber2023sources}, finding applications in areas such as Bayesian neural networks \citep{caprio_IBNN,kendall2017uncertainties}, adversarial attack detection mechanisms \citep{smith2018understanding}, and data augmentation strategies in Bayesian classification \citep{kapoor2022uncertainty}.
	
	Arguably, predictive uncertainty is the most studied form of uncertainty in both ML and statistics. It pertains prediction tasks such as those in supervised learning. 
	In the latter, we consider a hypotheses space $\mathcal{H}$, where each hypothesis $h \in \mathcal{H}$ maps a query instance $\boldsymbol{x}_q \in \mathcal{X}$ to a probability measure $p$ on $(\lab, \sigma(\lab))$, where $\lab$ denotes the outcome space, and $\sigma(\mathcal{Y})$ a suitable $\sigma$-algebra on $\lab$. By producing estimates of the ground-truth probability measure $p^*$ on $(\lab, \sigma(\lab))$, this probabilistic approach encapsulates aleatoric uncertainty about the actual outcome $y \in \mathcal{Y}$. Since epistemic uncertainty is difficult to represent with conventional probability distributions \citep{hullermeier2021aleatoric}, such predictions fail to capture the epistemic part of (predictive) uncertainty.  
	In order to account for both types of uncertainty,  machine learning methods founded on more general theories of probability such as imprecise probabilities or credal sets \citep{walley1991,augustin2014introduction} have been considered \citep{corani2012bayesian, caprio2023novel}.
	
	Another popular approach in this regard is to let the learner map a query instance $\boldsymbol{x}_q$ to a second-order distribution, i.e., a distribution on distributions, effectively assigning a probability to each candidate probability distribution $p.$
	Such an approach is realized, for example, by classical Bayesian inference \citep{gelman2013bayesian} or by the \emph{Evidential Deep Learning} (EDL) paradigm, which has recently become increasingly popular \citep{UlmerHF23}. 
	In the EDL paradigm, one essentially learns a model (usually a deep neural network) by empirical risk minimization, whose output for a query instance $\boldsymbol{x}_q$ are the parameters of a parameterized family of a second-order distribution. 
	So far, only the Dirichlet distribution has been used for classification, while the Normal-Inverse-Gamma distribution has been used for univariate regression \citep{amini2020deep} and the Normal-Inverse-Wishart distribution for multivariate regression \citep{malinin2020regression,meinert2021multivariate}. 
    However, this approach is not without controversy \citep{bengs_difficulty,meinert2023unreasonable,BengsHW23}.
	
	Regardless of the specific design of the EDL approach, the concrete quantification of the aleatoric $(\AU)$, epistemic $(\EU)$, as well as total uncertainty $(\TU)$ associated with the second-order predictive distribution plays a central role in any case.
	For regression, essentially, the variances on the different levels of the second-order distribution are used for this purpose, while measures from information theory are applied for classification: Shannon entropy	for $\TU$,  conditional entropy for $\AU$, and mutual information for $\EU$.
	Quite recently, \citet{wimmer2023quantifying} criticized the latter for not complying with properties that one could naturally expect of uncertainty measures for second-order distributions.
	However, the authors do not provide an alternative for reasonable quantification either, which, of course, would be of great importance for practical ML purposes, especially in safety-critical applications. 

	In this paper, we suggest an alternative way to obtain uncertainty measures in classification that overcome the drawbacks of the commonly used information-theory-based approach.
	To this end, we first propose a set of formal criteria that meaningful uncertainty measures for predictive uncertainty based on second-order distributions should obey. It extends the ones suggested by \cite{wimmer2023quantifying}. 
	Moreover, we provide a general framework based on distances on the second-order probability level for developing uncertainty measures to account for these criteria.
	Using the Wasserstein distance, we instantiate this framework explicitly and prove that all criteria are met.
	Finally, we elaborate on these quantities when the second-order distribution is a Dirichlet distribution.
	All proofs of the theoretical statements are provided in the supplementary material.
 
	\section{SECOND-ORDER UQ}
	\label{sec:uncertainty}
	In this section, we introduce the formal setting of supervised learning (throughout this paper we will exclusively deal with the case of classification) within which we establish further results. Let $(\mathcal{X}, \sigma(\mathcal{X}))$ and $(\mathcal{Y}, \sigma(\mathcal{Y}))$ be two measurable spaces. We will refer to $\mathcal{X}$ as \textit{instance} (or input) space and to $\mathcal{Y}$ as \textit{label} space, such that $|\mathcal{Y}| = K \in \mathbb{N}_{\geq 2}$. Further, we call the sequence $D = \{ ({x}_i, y_i )  \}_{i = 1}^n \in (\mathcal{X} \times \mathcal{Y})^n$ \textit{training data}. For $i \in \{1, \dots, n\}$, the pairs $(x_i, y_i)$ are realizations of random variables $(X_i, Y_i)$, which are independent and identically distributed (i.i.d.) according to some probability measure $p$ on $(\mathcal{X} \times \mathcal{Y}, \sigma(\mathcal{X} \times \mathcal{Y}))$. 
	Thus, each instance $x \in \mathcal{X}$ is associated with a conditional distribution $p(\cdot \, |\, x)$ on $(\lab, \sigma(\lab))$, such that $p(y \, | \, x)$ is the probability to observe label $y \in \mathcal{Y}$ given $x \in \mathcal{X}$.

	To ease the notation, we will denote by $\prob(\lab)$ the set of all probability measures on the measurable space $(\mathcal{Y}, \sigma(\mathcal{Y}))$. Similarly we write $\prob(\prob(\lab))$ for the set of all probability measures on $(\prob(\lab), \sigma(\prob(\lab))$; we refer to $Q \in \prob(\prob(\lab))$ as a \textit{second-order distribution}.\footnote{Note that there is no general consensus on terminology; thus, terms such as level-2 or type-2 distributions are also encountered in the literature.} 
	While usually upper-case letters denote probability measures and lower-case letters their pdf/pmf, in this paper we use capital letters for second-order and lower-case letters for fist-order distributions. 
	%
	The Dirac measure at $y \in \mathcal{Y}$ is denoted by $\delta_y \in \mathbb{P}(\mathcal{Y})$; likewise, $\delta_p \in \mathbb{P}(\mathbb{P}(\mathcal{Y}))$ denotes the Dirac measure at $p \in \mathbb{P}(\mathcal{Y}),$ where the underlying space of the Dirac measure should be clear from the context.
	Finally, $\mathrm{Unif}(\lab)$ denotes the uniform distribution on $\lab.$

\begin{figure}[ht!]
		\centering
		\includegraphics[width=.45\textwidth]{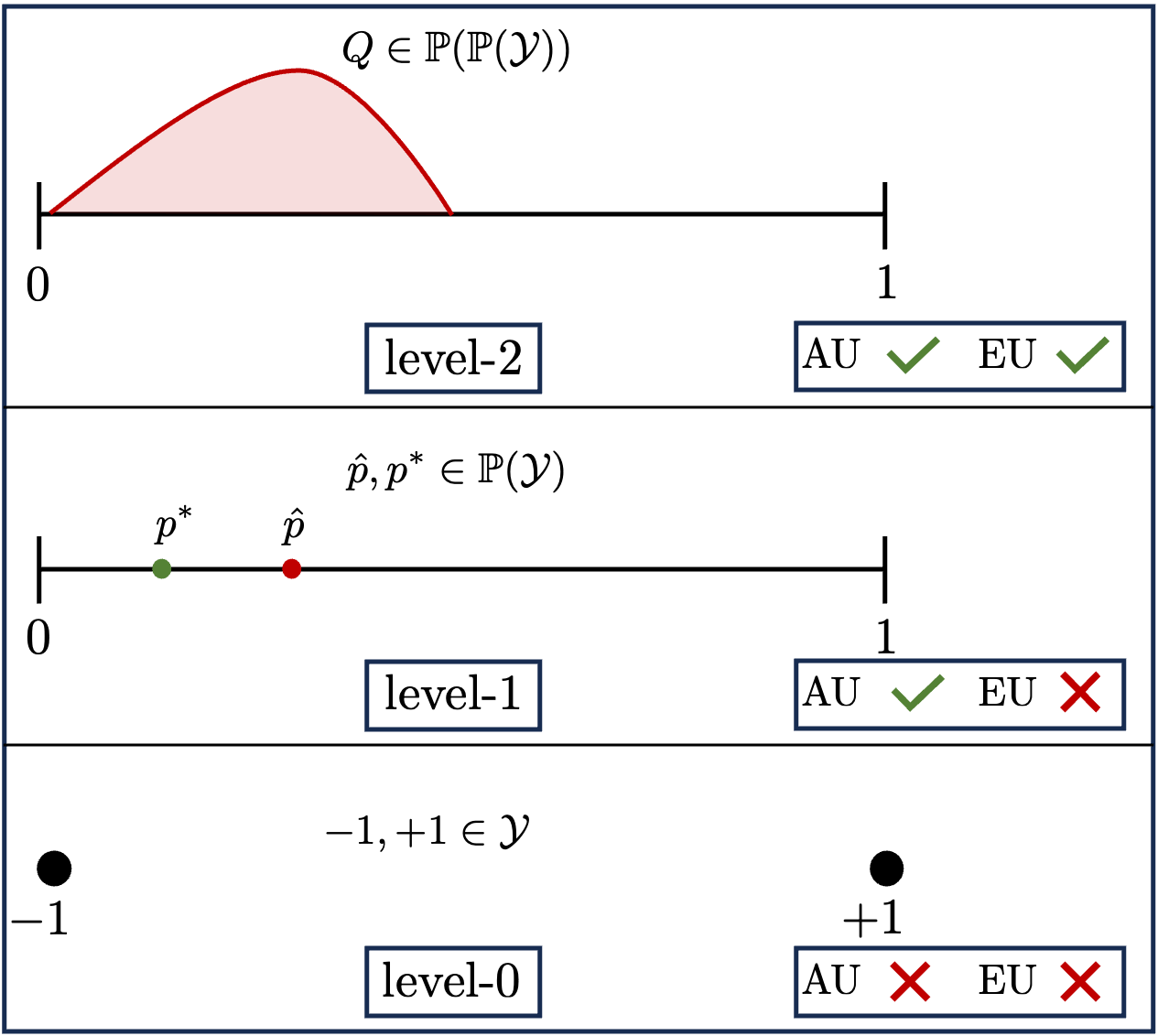}
		\caption{Uncertainty representation in binary classification, where $\mathcal{Y} = \{-1, +1\}$. From bottom to top increasing degrees of uncertainty awareness: Deterministic prediction, probabilistic prediction, and second-order prediction. See also \citep{wimmer2023quantifying}.}
		\label{fig1}
		\centering
	\end{figure}

	Given an instance $x \in \mathcal{X}$, let $Q \in \prob(\prob(\lab))$ denote the learner's current probabilistic belief\footnote{Although it would be more precise to let $Q$ depend on $x$, for ease of notation we will simply write $Q$.} about $p$, i.e., $Q(p)$ is the probability (density) of $p \in \prob(\lab)$. See Figure \ref{fig1} for an illustration of different levels (or degrees) of uncertainty-aware predictions.
	As already mentioned in the introduction, there are essentially two ways of obtaining such a second-order (predictive) distribution, either by means of Bayesian inference or via Evidential Deep Learning.
	Throughout the rest of this paper, we assume such a second-order predictive distribution $Q$ has been provided by a learner (though without being interested in how the prediction has been obtained).  
        We raise the question of how to quantify the total amount of uncertainty $(\TU)$, as well as the aleatoric $(\AU)$ and epistemic $(\EU)$ uncertainties associated with $Q$. 
	Moreover, we assume that our label space of interest $\mathcal{Y}$ is such that $| \mathcal{Y}| = K \in \mathbb{N}_{\geq 2}$.
 
	\subsection{Default Measures of Uncertainty}
	\label{sec:related}
	We begin by revisiting the arguably most common information-theoretic approach in machine learning for measuring predictive uncertainty in classification tasks. 
	This approach exploits (Shannon) \textit{entropy} and its link to  mutual information and conditional entropy for   specifying explicit quantities for the total ($\TU$), aleatoric ($\AU$), and epistemic ($\EU$) uncertainties associated with a predictive second-order distributions $Q \in \pplab$ 
\citep{houlsby_2011_BayesianActiveLearning,gal_2016_UncertaintyDeepLearning,depeweg2018decomposition, smith2018understanding, mobiny_2021_DropConnectEffectiveModeling}. 

	The (Shannon) \textit{entropy} \citep{shannon1948mathematical} of $p \in \plab$ is defined as
	\begin{align}
		H(p) \coloneqq - \sum\nolimits_{y \in \lab} p(y) \log_2 p(y).
		\label{eq:entropy}
	\end{align}
	We can analogously define the entropy of a (discrete) random variable  $Y: \Omega \longrightarrow \lab$ by
	\begin{align}
		H(Y) \coloneqq - \sum\nolimits_{y \in \lab} p_Y(y) \log_2 p_Y(y),
	\end{align}
	where $p_Y \in \plab$ is the corresponding push-forward measure on the measurable space $(\lab,  2^{\lab})$.
    The Shannon entropy has established itself as a standard measure of uncertainty due to its appealing theoretical properties and intuitive interpretation. Specifically, it measures the degree of uniformity of the distribution $p_Y$ of a random variable $Y$, and corresponds to the log-loss of $p_Y$ as a prediction of $Y$.
    
	In the following, we assume $p_R \sim Q,$ i.e., $p_{R}: \Omega^{\prime} \rightarrow \plab$ is a random first-order distribution distributed according to a second-order distribution $Q$ and consequently taking values in the $(K-1)$-dimensional probability simplex. 
	For $\omega^{\prime} \in \Omega^{\prime}$, we denote by $p = p_{R}(\omega^{\prime})$ the realization of  $p_{R},$ respectively.  
	
	The core idea for obtaining uncertainty measures for a given second-order distribution $Q$ is to consider the expectation of $Q$ given by 
	\begin{align}
		\overline{p} = \mathbb{E}_Q[p_R] = \int_{ \mathbb{P}(\mathcal{Y})} p \, \mathrm{d}Q(p) \, , 
		\label{eq:aggregation}
	\end{align} 
	which yields a probability measure $\overline{p}$ on $(\mathcal{Y}, \sigma(\mathcal{Y})),$ i.e., a first-order distribution. 
 
	With this, it seems natural to define the measure of total uncertainty as the entropy \eqref{eq:entropy} of $\overline{p} \in \plab$.
	More precisely, total uncertainty associated with a second-order distribution $Q \in \pplab$ can be computed as 
	\begin{align}
		\TU(Q) = H\left( \mathbb{E}_Q[p_R]   \right). 
		\label{tu:entropy}
	\end{align}
	In a similar fashion, one defines aleatoric uncertainty as \textit{conditional entropy} 
	\begin{align}
		\AU(Q) = \mathbb{E}_Q[ H(Y|p_{R}) ] = \int_{\plab} H(p) \; \mathrm{d}Q(p). 
		\label{au:entropy}
	\end{align}
	
	Further, the measure of epistemic uncertainty is in particular motivated by the well-known additive decomposition  of entropy into conditional entropy and \textit{mutual information} 
    \citep[Equation (2.40)]{cover1999elements}, i.e.,
	\begin{align}
		H(Y) = H(Y\, | \, p_{R})+ I(Y,p_{R}). 
		\label{eu:entropy}
	\end{align}
	By rearranging \eqref{eu:entropy} we get a measure of epistemic uncertainty
	\begin{align} \label{eu:entropy_2}
		\begin{split}
			\EU(Q) = I(Y,p_{R}) &= H(Y)- H(Y\, | \, p_{R}) \\[0.2cm]
			&= \mathbb{E}_Q[D_{KL}(p_{R} \, \| \, \overline{p})],
		\end{split}
	\end{align}
	where $D_{KL}(\cdot \, \| \, \cdot)$ denotes the Kullback-Leibler (KL) divergence (or distance) \citep{kullback1951information}. 
 
    Even though the individual measures, i.e., entropy, mutual information, and conditional entropy, have reasonable interpretations in terms of quantifying the respective uncertainty, which are particularly useful when applied to first-order predictive distributions, a different picture emerges for the above approach to second-order predictive distributions.
	Some issues regarding the quantification of the respective uncertainties have recently been intensively discussed by \cite{wimmer2023quantifying}, which we will take up and elaborate on in the following section. 
	Essentially, the problem stems from $\TU$ in \eqref{tu:entropy} and $\EU$ in \eqref{eu:entropy_2} depending on the second-order predictive distribution $Q$ only through their expectation  $\overline{p}$ in \eqref{eq:aggregation}.

    \subsection{Alternatives for the Default Measures}
    Recently, a variant of the above approach was proposed, which attempts to overcome the issues mentioned \citep{schweighofer2023introducing}.
    For this purpose, the total uncertainty in \eqref{tu:entropy} is rewritten as
    \begin{align*}
		\TU(Q) =  \mathbb{E}_Q[ CE(p_R,\tilde p)],
    \end{align*}
    where $CE(\cdot,\cdot)$ is the cross-entropy, i.e.,
    \begin{align*}
        CE(p,q) \coloneqq - \sum\nolimits_{y \in \lab} p(y) \log_2 q(y)
    \end{align*}
    for $p,q \in \mathbb{P}(\lab)$. Then, the  alternative measure for total uncertainty suggested by the authors is  
    \begin{align}
		\TU(Q) =  \mathbb{E}_{Q,Q^{\prime}}[ CE(p_R, p_R^{\prime})],
		\label{tu:entropy_improved}
    \end{align}
    where $Q^{\prime}$ is an i.i.d.\ copy of $Q$ and $p_R^{\prime} \sim  Q^{\prime}.$ 
    Using again the decomposition in \eqref{eu:entropy} and the resulting components as measures for aleatoric and epistemic uncertainty, one obtains the same aleatoric uncertainty measure as in \eqref{au:entropy}, but the epistemic uncertainty measure changes to
    \begin{align} \label{eu:entropy_improved}
        \EU(Q) = \mathbb{E}_{Q,Q^{\prime}}[ D_{KL}(p_R \, \| \, p_R^{\prime})].
    \end{align}

        \section{Novel Uncertainty Meausures}
	\subsection{Axiomatic Foundations}
	\label{subsec:axiom}
	
	The criticism of the previous approach raised by \cite{wimmer2023quantifying} is grounded in the postulation of criteria that measures of total, aleatoric, and epistemic uncertainties should naturally satisfy when used for quantifying predictive uncertainty associated with second-order distributions. 
    This is similar to the literature on uncertainty quantification for other methods of representing uncertainty, such as belief functions or credal sets \citep{bronevich2008axioms, pal1993uncertainty, sale2023volume}. 
	In the following, we build on -- and extend -- the criteria presented by \cite{wimmer2023quantifying}.
 
	For this purpose, we let $\TU$, $\AU$, and $\EU$ denote, respectively, measures $\prob(\prob(\lab)) \to \mathbb{R}_{\geq 0}$ of total, aleatoric, and epistemic uncertainties associated with a second-order uncertainty representation $Q \in \prob(\prob(\lab))$.
	If $\lab_1$ and $\lab_2$ are partitions of $\lab$ and $Q \in \prob(\prob(\lab))$, then we denote by $Q_{|\lab_i}$  the marginalized distribution on $\lab_i$. 
	In the same spirit, we define $\TU_{\lab_i}$.

	\begin{enumerate}[noitemsep,topsep=0pt,leftmargin=6mm]
		\item[A0] $\TU$, $\AU$, and $\EU$ are non-negative. 
		\item[A1] $\AU(\delta_{\mathrm{Unif}(\lab)}) \geq \AU(\delta_p) \geq \AU(\delta_{\delta_{y}}) = 0$ holds for any $ y  \in \lab$  and any  $p \in \plab.$ 
		\item[A2] $\EU(Q) \geq \EU(\delta_p) = 0$ holds for any $Q \in \pplab,$ and any $p \in \plab$. Further, for any $Q \in \pplab$ with $\AU(Q) = 0$ we have $\EU(Q') \geq \EU(Q)$, where $Q^{\prime}$ is such that $Q^{\prime}(\delta_y) = \frac{1}{K}$ for all $y \in \lab$. 

        \item[A3] $\AU(Q) \leq \TU(Q)$ and $\EU(Q) \leq \TU(Q)$  holds for any $Q \in \pplab.$
		\item[A4] $\TU(Q)$ is maximal for $Q$ being the continuous second-order uniform distribution.
	\item[A5] If $Q'$ is a mean-preserving spread of $Q$,\footnote{Let $X \sim Q, X^\prime \sim Q^\prime$ be two random variables, where $Q, Q' \in \prob(\prob(\lab))$. Then, $Q^\prime$ is called a mean-preserving spread of $Q$ iff $X^\prime \overset{d}{=} X + Z$, for some random variable $Z$ with $\mathbb{E}[Z| X = x] = 0$, for all $x$ in the support of $X$.} then $\text{$\EU$}(Q') \geq \text{$\EU$}(Q)$ (weak version) or $\text{$\EU$}(Q') > \text{$\EU$}(Q)$ (strict version). 
        \item[A6] If $Q'$ is a spread-preserving location shift of $Q$,\footnote{
			$Q$ and $Q^\prime$ differ only in their respective means.} then $\text{$\EU$}(Q') = \text{$\EU$}(Q)$.
		\item[A7] $\TU_{\lab}(Q) \leq \TU_{\lab_1}(Q_{|\lab_1}) + \TU_{\lab_2}(Q_{|\lab_2}).$
		\item[A8] $\TU_{\lab}(Q_{|\lab_1} \otimes    Q_{|\lab_2}) = \TU_{\lab_1}(Q_{|\lab_1}) + \TU_{\lab_2}(Q_{|\lab_2})$, where $\otimes$ denotes the product measure \citep{durrett2010probability}.
	\end{enumerate}
	Before discussing each criterion\footnote{A0 is a trivial property and therefore not discussed. }, we first start with a joint and more in-depth discussion of A1 and A2, since they play a central role in the discussion of most of the other criteria.

	\textbf{A1 and A2:} Since we are interested in second-order distributions $Q$ for the purpose of predictive uncertainty, it is natural to speak of a state of absence of  epistemic uncertainty if $Q$ correspond to a point mass on the second-level.
	This is reflected by the lower bound in A2 and is also a viewpoint shared in the literature \citep{bengs_difficulty,wimmer2023quantifying}.
	Moreover, there is agreement in the literature that (i) the uniform distribution on the first level, i.e.\ $\mathrm{Unif}(\mathcal{Y}),$ represents the case of highest outcome uncertainty, (ii) a degenerated first-order distribution, i.e., a Dirac measure on a point $y\in \mathcal{Y},$ represents the case of lowest outcome uncertainty, and (iii) first-order distributions between these extreme cases correspond to an outcome uncertainty that lays somewhere ``in-between''.
	In the absence of epistemic uncertainty in the predictive second-order distribution, this should be reflected by the measure of aleatoric uncertainty ($\leadsto$ A1).

    If the uncertainty is only epistemic in nature, that is, if according to A1 only first-order Dirac measures remain as possible candidates, then the epistemic uncertainty should be maximal when the ambiguity around the Diracs is maximal. This happens when the second-order distribution $Q$ is a discrete uniform distribution on the first-order Dirac measures on the elements of $\mathcal{Y}$ ($\leadsto$ A2).
	Note that this view differs from that of \cite{wimmer2023quantifying}, which demands maximum epistemic and total uncertainties for the continuous second-order uniform distribution. 
	However, our criteria are consistent regarding the maximal total uncertainty ($\leadsto$ A4). 
	
	\textbf{A3:} As discussed in detail by \citet[Section 4.4]{wimmer2023quantifying}, the aleatoric and epistemic uncertainties of a second-order predictive distribution are closely intertwined.
    Since total uncertainty subsumes both types of uncertainty simultaneously, it should be always an upper bound for $\AU$ and $\EU$, respectively.

	\textbf{A5 and A6:} These properties are again inspired by \citep{wimmer2023quantifying}. If two second-order distributions have the same expectation (i.e., ``agree'' on the aleatoric uncertainty) but differ in their dispersion or spread, the distribution with higher dispersion should be assigned higher epistemic uncertainty ($\leadsto$ A5).  
    Similarly, with equal dispersion, epistemic uncertainty should be the same in all cases. Thus, if $Q \in \pplab$ and $Q^{\prime} \in \pplab$ only differ in their respective means, epistemic uncertainty should be the same in both cases ($\leadsto$ A6).
    
	\textbf{A7 and A8:} These criteria are inspired by the those underlying Shannon entropy.
	Specifically, these properties aim to ensure that the total uncertainty of a second-order predictive distribution does not exceed the total uncertainties over all possible marginalizations of it with respect to the label space $\mathcal{Y}.$ 
	Thus, a subadditivity property should also hold here ($\leadsto$ A7), with equality achieved when the marginalizations are independent ($\leadsto$ A8).

	As shown by \citet{wimmer2023quantifying}, the measures for total, aleatoric, and epistemic  uncertainties in (\ref{tu:entropy}-\ref{eu:entropy_2})  fail to satisfy A5 and A6 when it comes to second-order distributions.
    For the alternative version of these measures suggested by \citet{schweighofer2023introducing} it is not shown whether these properties are fulfilled or not. 
    However, total uncertainty in \eqref{tu:entropy_improved} will not be maximal for $Q$ being the continuous second-order distribution, but for $Q'$ as in A2, so violating A4.
    In addition, it is apparent from the definition that both $\TU$ and $\EU$ in \eqref{tu:entropy_improved} and \eqref{eu:entropy_improved} can go to infinity.  
    Thus, the measures are not naturally restricted to an interpretable range.
    %

        \subsection{Distance-based Measures}
	\label{sec:proposal}
	
	We now introduce a general framework for deriving suitable measures for aleatoric, epistemic, and total uncertainties based on a second-order  distribution $Q \in \prob(\prob(\lab))$.
	The main constituents of the framework are (i) a (suitable) distance $d_2(\cdot, \cdot)$ on $\pplab$ and (ii) specific reference sets of second-order distributions representative for $\AU,\EU$ or $\TU$, respectively, each lacking one or both types of uncertainties. 
	Roughly speaking, each uncertainty measure (i.e., $\AU,\EU$ or $\TU$) of $Q$ is defined as the minimal distance of $Q$ to the corresponding reference set.
	This approach is inspired by the field of \textit{optimal transport} \citep{villani2009optimal, villani2021topics} and guided by the following question:
	``How much do we need to move $Q$ to arrive at the nearest second-order distribution of the respective reference set for $\AU,\EU$ or $\TU$?''
	While the distance function -- according to which $Q$ moves in the space $\mathbb{P}(\mathbb{P}(\mathcal{Y}))$ -- is intentionally kept flexible in our framework, the reference sets are fixed 
    and should naturally lead to the fulfillment of A0--A8, ideally for a broad class of distances. 
	
	\paragraph{Total uncertainty ($\TU$).} For the total uncertainty we suggest to use all second-order Dirac measures on the set of first-order Dirac measures as the reference set.
	More specifically, total uncertainty is defined as 
	\begin{align}
		\TU(Q) \coloneqq \min\nolimits_{y \in \lab} \, d_2(Q, \delta_{\delta_{y}}).
		\label{tu:new}
	\end{align}
	This choice of the reference set is natural as each element in this reference set represents the case of an absolutely certain prediction/decision, i.e., there is neither aleatoric (first-order) nor epistemic (second-order) uncertainty present.
	Thus, the farther $Q$ is from such an element, the farther one is from making a decision without any kind of uncertainty, which is reflected by \eqref{tu:new}.
 
	\paragraph{Aleatoric uncertainty ($\AU$).} The reference set for aleatoric uncertainty should be the set of all mixtures of second-order Dirac measures on first-order Dirac measures, i.e., 
 \begin{align*} 
\Delta_{\delta_m} = \Big\{ \delta_m \in & \prob(\prob(\lab)): \\
& \delta_m =   \sum_{y \in \lab} \lambda_y \cdot \delta_{\delta_{y}}, \, \sum_{y \in \lab} \lambda_y = 1 \Big\} \, .
 \end{align*}
	If we agree on A0--A8, each element in this set has no aleatoric uncertainty, so the assessment of a second-order distribution $Q$ is solely in terms of its amount of aleatoric uncertainty.
	Accordingly, the measure of aleatoric uncertainty is defined as 
	\begin{align}
		\AU(Q) \coloneqq \min\nolimits_{\delta_m \in \Delta_{\delta_m}} d_2(Q, \delta_m).
		\label{au:new}
	\end{align}
	
	\paragraph{Epistemic uncertainty ($\EU$).} In the same spirit as \eqref{au:new}, we want to assess $Q$ solely in terms of its amount of epistemic uncertainty.
	Again by agreeing on A0--A8, we naturally obtain as reference set the collection of all second-order Dirac measures on the probability simplex, since these have no epistemic uncertainty. 
    If we denote the latter by $\Delta_{\delta_p}$, we obtain for the measure of epistemic uncertainty 
	\begin{align}
		\EU(Q) \coloneqq \min\nolimits_{\delta_p \in \Delta_{\delta_p}} d_2(Q, \delta_p).
		\label{eu:new}
	\end{align} 
	It is worth noting that the entropy-based uncertainty measures in Section \ref{sec:related} can also be considered from the perspective of our distance-based framework.
	Indeed, the entropy of a (discrete) distribution is related to the negative KL divergence (or KL distance) between the distribution and the uniform distribution (on the respective domain) \citep[Equation (2.107)]{cover1999elements}. 
    Thus, we could rewrite 	\eqref{tu:entropy}, \eqref{au:entropy} and \eqref{eu:entropy_2} as 
	\begin{align}
		\begin{split}
			\TU(Q) &= \log K - D_{KL}( \mathbb{E}_Q[p_R]  \, \| \,\mathrm{Unif}(\lab)), \\
			\AU(Q) &=  \log K  - \mathbb{E}_Q[ D_{KL}( p_{R} \, \| \, \mathrm{Unif}(\lab)) ], \\
			\EU(Q) &= \mathbb{E}_Q[D_{KL}(p_{R} \, \| \, \overline{p})].
		\end{split}
			\label{eq:reframe_entropy}
	\end{align}
        With this representation, we see that the $\EU$ measure \eqref{eu:entropy_2} has similarities to ours. More specifically, it is obtained as a special case of \eqref{eu:new} with $d_2(\cdot , \cdot)$ the expected KL divergence (for which the minimum is obtained by $\delta_p = \bar{p}$). Note, however, that the expected KL divergence is not a proper distance on $\pplab$, wherefore \eqref{eu:entropy_2} is not a special case of our framework in a strict sense. 
        Moreover, the interpretation of the measures $\TU$ and $\AU$ is different from our measures \eqref{tu:new} and \eqref{au:new},
        as both are measuring \emph{similarity} (through the negated KL divergence) to the case of maximal uncertainty, namely the first-level uniform distribution, instead of dissimilarity to a reference set of least uncertain distributions.

    The alternative version (\ref{tu:entropy_improved}--\ref{eu:entropy_improved}) suggested by \citet{schweighofer2023introducing} does not have such an interpretation, except for the aleatoric uncertainty which remains the same. 
    This is due to the lack of a reference set for $\TU$ and $\EU,$ so that both measures are more interpretable as a measure of the diversity of the second-order distribution. 
    On a high level, the approach also follows the idea of including the entire characteristics (first \emph{and} second level) of $Q$ in the respective uncertainty assessment, instead of narrowing down to the expected value $\overline{p}$ in \eqref{eq:aggregation} like the default case. 
 
    %
     
	\section{WASSERSTEIN AS DISTANCE}
        \subsection{General Case}
	
	So far, we did not specify the distance $d_2: \prob(\prob(\lab)) \times \prob(\prob(\lab)) \rightarrow \mathbb{R}_{\geq 0}$ on $\prob(\prob(\lab))$. In the following, we will motivate one specific choice, namely
	the \textit{Wasserstein distance} (or Kantorovich--Rubinstein metric). For our discussion, we first recall the concept of \textit{coupling}, a term that is central to optimal transport theory \citep[Chapter 1]{villani2009optimal}. 
	Note that the definition used in this paper is an adaptation of the standard one, as our focus is on second-order distributions.
	
	\begin{defi} \label{def:Coupling}
		We call the probability measure $\gamma$ on $(\plab \times \plab, \sigma(\plab \times \plab)$ coupling of $P, Q \in \pplab$ iff for all measurable sets $A, B \in \sigma(\plab)$ one has 
		$$\gamma[A \times \plab] = Q[A],\ \mbox{and} \ \gamma[\plab \times B] = P[B].$$
		Thus, $\gamma$ admits marginals $P$ and $Q$.
	\end{defi}
	
	Let $(\prob(\lab), d_1)$ be a metric space, where $\lab$ is defined as before, and $d_1$ is a suitable metric on the space $\prob(\lab)$ (again, equipped with a suitable $\sigma$-algebra). Then, for $p\in[1,\infty]$ the (second-order) $p$-Wasserstein distance between two probability measures $P, Q \in \prob(\prob(\lab))$ is defined as
	\begin{align}
		\mathcal{W}_p(P, Q) = \left( \inf_{\gamma \in \Gamma(P, Q)  } \int_{\prob(\lab) \times \prob(\lab)} d_1(p, q)^p \, \mathrm{d}\gamma (p, q)  \right)^{\tfrac{1}{p}}
		\label{distance:wp2}
	\end{align}
	where $\Gamma(P, Q)$ denotes the set of all couplings between the probability measures $P$ and $Q$ (see Definition \ref{def:Coupling}). 

	The choice of this metric for our purposes is quite natural based on its interpretation: The Wasserstein metric quantifies how much mass has to be moved around and how far in order to convert one distribution into another. 
	This is perfectly in line with our view for the uncertainty measures in Section \ref{sec:proposal}.  
	In accordance with the literature, we will be exclusively concerned with the case $p = 1$ and omit in the following the subscript in $\mathcal{W}_p(\cdot, \cdot)$. First, we show that $\mathcal{W}(\cdot, \cdot)$ is indeed a well-defined metric on $\pplab$.
 
	\begin{lemma}\label{lemma_wassers}
		 The second-order Wasserstein distance $\mathcal{W}: \mathbb{P}(\mathbb{P}(\mathcal{Y})) \times \mathbb{P}(\mathbb{P}(\mathcal{Y})) \rightarrow \mathbb{R}_{\geq 0}$ is a well-defined metric on $\pplab$.
	\end{lemma}
	
	Since in both \eqref{tu:new} and \eqref{eu:new} second-order Dirac measures are involved, we show now that the optimal coupling between a second-order distribution $Q \in \pplab$ and a second-order Dirac measure $\delta_p$, where $p \in \plab$, is trivially given by the respective product measure. This simplifies corresponding computations. 
	
	\begin{prop} \label{prop_optimal_coupling}
		For any second-order Dirac measure $\delta_p \in \prob(\prob(\lab))$, $p \in \prob(\lab)$, and any second-order distribution $Q \in \prob(\prob(\lab))$, the optimal coupling between $\delta_p$ and $Q$ is the product measure $\gamma = Q \otimes \delta_p$.
	\end{prop}
	
	This coupling is also frequently referred to as \textit{trivial coupling} \citep{villani2009optimal}.
 	Let us elaborate on the choice of the metric $d_1: \prob(\lab) \times \prob(\lab) \rightarrow \mathbb{R}_{\geq 0}$ in \eqref{distance:wp2}. We will define this as the Wasserstein metric between two first-order distributions induced by the \textit{trivial} distance on the label space $\lab$. Note that this is not fixed by design, and without loss of generality other metrics on the label space (depending on the specific problem at hand) can be considered. The trivial distance on  $\lab$ is given for any $y, y' \in \lab$ by
	\begin{align}
		d_0(y, y') = \left\{\begin{array}{lr}
			1, & \text{for } y \neq y' \\
			0, & \text{for } y = y' 
		\end{array}\right.  \, .
		\label{distance:trivial}
	\end{align}

	With the choice of the distance \eqref{distance:trivial}, for $p, q \in \prob(\lab)$ we obtain the following \textit{induced} first-order distance $d_1$:
	\begin{align}
		d_1(p,q) &= W(p,q) \nonumber \\[0.2cm]
		&= \inf_{\gamma \in \Gamma(p, q)} \int_{\lab \times \lab} d_0(y,y') \, \mathrm{d}\gamma(y,y') \nonumber \\[0.2cm]
		&= \inf_{\gamma\in \Gamma(p, q)} \sum_{y,y' \in \lab}  d_0(y,y') \, \gamma(y,y') \nonumber  \\[0.2cm]
		&= \inf_{\gamma\in \Gamma(p, q)} \sum_{\substack{y, y' \in \lab \nonumber \\ y \neq y'} }\gamma(y,y') \nonumber \\[0.2cm]
		&= \inf_{\gamma\in \Gamma(p, q)} \left\{ 1- \sum_{y \in \lab} \gamma(y,y) \right\} .
		\label{wasserstein01:min}
	\end{align}

	Note the nuance in our notation. We use calligraphic $\mathcal{W}(\cdot, \cdot)$ for the second-order Wasserstein distance, and $W(\cdot, \cdot)$ for the first-order Wasserstein distance. 
	
	\begin{prop}
		\label{prop:min}
		The coupling $\gamma \in \Gamma(p, q)$ minimizing the expression \eqref{wasserstein01:min} is such that $\gamma(y,y) = \min\{p(y), q(y)\}$.    
	\end{prop}
	
	Proposition \ref{prop:min} yields
	\begin{align*}
		d_1(p,q) &= 1- \sum_{y \in \lab} \min\{p(y), q(y)\} \\[0.2cm]
		&= \frac{1}{2}\sum_{y \in \lab} \max\{p(y), q(y)\} - \min\{p(y), q(y)\} \\[0.2cm]
		&= \frac{1}{2}\sum_{y \in \lab} |p(y) - q(y)| \\[0.2cm]
		&= \frac{1}{2} \lVert p - q \rVert_1. 
	\end{align*}
    In the context of usual probability measures, Proposition \ref{prop:min} is well-known in transportation theory, establishing a connection between the Wasserstein metric and the Total Variation Distance (TVD). 
    With this, the proposed measures for $\TU$, $\AU$, and $\EU$ for $Q \in \prob(\prob(\lab))$ in Section \ref{sec:proposal} simplify as follows.
	
	\begin{prop} \label{prop:explicit_repr}
		With the above choice of $d_1(\cdot, \cdot)$ as distance on $\plab$, previously defined measures of uncertainty in \eqref{tu:new}, \eqref{au:new} and \eqref{eu:new} simplify to 
		\begin{align}
			\TU(Q) &=1-\max\nolimits_{y \in \lab } \mathbb{E}_{Q_y}[p(y)], \label{tu:simple}\\[0.1cm]
			\AU(Q) &= 1- \mathbb{E}_Q \left[\max\nolimits_{y \in \lab } p(y) \right], \label{au:simple}\\[0.1cm]
			\EU(Q) &= \frac{1}{2} \min\nolimits_{q \in \prob(\lab)} \mathbb{E}_{p \sim Q} [\lVert p - q \rVert_1].
			\label{eu:simple}
		\end{align}
		Here, $Q_y$ denotes the marginal distribution associated with $Q \in \prob(\prob(\lab)$ for some  $y \in \lab.$
	\end{prop}
	
	The following proposition elaborates on the ranges of the proposed measures of uncertainty. While the results appear natural, they yield interesting findings from an uncertainty quantification perspective.
	
	\begin{prop} \label{prop:ranges}
		With the choice of $d_1(\cdot, \cdot)$ as distance on $\plab$, we have the following:
		\begin{itemize}[noitemsep,topsep=0pt,leftmargin=8mm]
			\item[i.)] $\forall Q \in \pplab: \TU(Q) \leq \frac{K-1}{K}$, where the upper bound is reached for $Q^{\prime} \in \pplab$ such that $\mathbb{E}_{Q^{\prime}}[p] = \mathrm{Unif}(\lab)$.
			\item[ii.)] $\forall Q \in \pplab: \AU(Q) \leq \frac{K-1}{K}$, where the upper bound is reached for $Q^{\prime} = \delta_{\mathrm{Unif}(\lab)}$. 
			\item[iii.)] $\forall Q \in \pplab: \EU(Q) \leq \frac{K-1}{K}$, where the upper bound is reached for any $Q^{\prime} \in \pplab$  such that $Q^{\prime}(\delta_y) = \frac{1}{K}$ for all $y \in \lab$.
		\end{itemize}
	\end{prop}
	The property from Proposition \ref{prop:ranges} is desirable for two reasons. On the one hand, the value range grows with increasing complexity of the classification problem in terms of the number of labels $K$. 
	This is similar to the entropy, see \eqref{eq:reframe_entropy}. 
	On the other hand, the value ranges are ``normalizing themselves'' with increasing complexity.  
	More precisely, for $K \rightarrow \infty$, the maximum of the uncertainty measures converges (with respect to the standard Euclidean metric on $\mathbb{R}$) to $1$. 
	Needless to say, the upper bounds of the value ranges can also be used to normalize the uncertainty measures a priori by multiplying them with $K/(K-1).$  
 
    A direct consequence of Proposition \ref{prop:ranges} is that maximum epistemic uncertainty can be achieved only when there is no aleatoric uncertainty and vice versa. 
    \begin{corollary}
    \label{corallary:31}
    For any $Q \in \pplab$, it holds that $\EU(Q) = \frac{K}{K-1}$ if and only if $\AU(Q) = 0.$ 
    \end{corollary}
	Finally, we show that the proposed uncertainty measures with the Wasserstein distance instantiations fulfill the criteria specified in 
	Section \ref{subsec:axiom}.
	
	\begin{theorem} \label{theorem:axioms}
		The uncertainty measures (\ref{tu:new}-\ref{eu:new}) with the Wasserstein distance instantiation satisfy A0--A8 as defined in Section \ref{subsec:axiom}.
	\end{theorem}
	
        \subsection{Dirichlet Distribution}
	Owing to its key role as a conjugate prior for a categorical distribution, the Dirichlet distribution is arguably the most important family of parameterized  (second-order) distributions employed in various areas of theoretical and applied research. 
	In Bayesian inference and Evidential Deep Learning, the Dirichlet distribution has become the gold standard.   
	Accordingly, in this section we focus on the computation of the proposed uncertainty measures with the Wasserstein distance initialization for the case of Dirichlet distributions. 
	We start with a brief introduction to the Dirichlet distribution and identify w.l.o.g.\ each element in the label space $\lab$ with an integer, i.e.\ $\lab = \{1,2,\ldots,K\}.$ 

	Let $\pi$ denote a $K$-dimensional probability vector, and assume it is distributed according to a Dirichlet distribution, that is, $\pi \sim \text{Dir}(\bm{\alpha})$. The \textit{Dirichlet distribution} $\text{Dir}(\bm{\alpha})$ is supported on the $(K-1)$-dimensional unit simplex, and it is parameterized by $\bm{\alpha}=(a_1,\ldots,a_K)^\top$, a $K$-dimensional vector whose entries are such that $\alpha_j > 0$, for all $j\in\{1, \ldots, K\}$. Its probability density function (pdf) is given by 
	\begin{center}
		$\frac{1}{B(\bm{\alpha})} \cdot \prod_{j=1}^K \pi_j^{\alpha_j-1},$
	\end{center}
	 where $B(\cdot)$ denotes the multivariate Beta function. We can interpret the $j$-th entry $\alpha_j$ of $\bm{\alpha}$ as pseudo-counts: $\alpha_j$ represents the virtual observations that we have for label $j$. It captures the agent's (i.e., machine learning algorithm's) knowledge around label $j$ that comes e.g. from previous or similar experiments. The expected value of $\pi \sim \text{Dir}(\bm{\alpha})$ is given by $\mathbb{E}(\pi_j)=\alpha_j/\sum_j \alpha_j$, $j\in\{1,\ldots,K\}$, and it expresses the belief that $j$ is the ``true label''. 
	This is due to the fact that the marginals $\pi_j$ of the Dirichlet distribution are distributed according to a Beta distribution with parameters $\alpha_j$ and $\alpha_0 - \alpha_j$, with $\alpha_0 = \sum_{i=1}^{K} \alpha_i$. 
	 
	Dirichlet distributions are second-order distributions since their support is the $(K-1)$-dimensional simplex, i.e.\ $\plab$. That is, they can be thought of as distributions over the actual probability measures that generated the data.
 
	In the following,  we assume that our current probabilistic knowledge is given by $Q \sim \Dir(\bm{\alpha})$, so that the marginal distributions are Beta distributions, i.e., $Q_i \sim \text{Beta}(\alpha_i, \alpha_0 - \alpha_i)$ with $\alpha_0 = \sum_{j=1}^{K} \alpha_j$ for each $i \in \lab$. 
	Using the closed-form for the expectation of the marginals, we immediately obtain 
	\begin{align}
		\TU(Q)  = 1- \max\nolimits_{y \in \lab } \frac{\alpha_y}{\alpha_0}
	\end{align} 
    for the total uncertainty in \eqref{tu:simple}. Unfortunately, it is difficult to derive a closed form for the expression in \eqref{au:simple}. 
    However, the expected value is easily approachable through Monte Carlo simulations. 
	
	Finally, for $\EU$ in \eqref{eu:simple}, we are dealing with a constrained optimization problem, which, however, has appealing properties.
	Indeed, given a Dirichlet distribution, we seek to solve the following constraint optimization problem for \eqref{eu:simple}:
	\begin{align}
		&\underset{q \in (0,1)^K}{\textbf{minimize}} \quad h(q) \coloneqq \frac{1}{2} \sum\nolimits_{i=1}^{K} \mathbb{E}_{p_i \sim Q_i} [\, | p_i - q_i | \,] \label{opt:objm} \\[0.3cm]
		&\textbf{subject to} \quad c(q) \coloneqq \sum\nolimits_{i = 1}^{K} q_i - 1 = 0
		\label{opt:constm}
	\end{align}

	Further evaluation of the sum of expectations involved in \eqref{opt:objm} yields
\begin{sizeddisplay}{\small}
	\begin{align}
		\frac{1}{2} \sum_{i=1}^{K} \left[\frac{\alpha_i}{\alpha_0}(1-2F_{\alpha_i+1, \alpha_0 - \alpha_i}(q_i)) + q_i \left(2F_{\alpha_i,\alpha_0 - \alpha_i}(q_i) - 1 \right) \right]
		\label{eq:hp}
	\end{align}
 \end{sizeddisplay}
	where $F_{\alpha,\beta}(\cdot)$ denotes the cdf of $\text{Beta}(\alpha,\beta)$. Consequently, the \textit{Lagrangian function} is given by
	\begin{align}
		\mathcal{L}(q, \lambda) &= h(q) + \lambda \, c(q),
		\label{eq:lag}
	\end{align}
	where $\lambda \in \mathbb{R}$. The extreme points of the Lagrangian \eqref{eq:lag} are the solutions of the equations 
	\begin{align}
		\nabla h(q) = -\lambda \nabla c(q), \quad 
		\sum\nolimits_{i = 1}^{K} q_i - 1 = 0.
	\end{align}
	
	\begin{prop} \label{prop:eu_dirichlet}
		The constraint optimization problem (\ref{opt:objm}--\ref{opt:constm}) has a unique solution. 
	\end{prop}

    \begin{figure}[ht!]
		\centering
		\includegraphics[width=.5\textwidth]{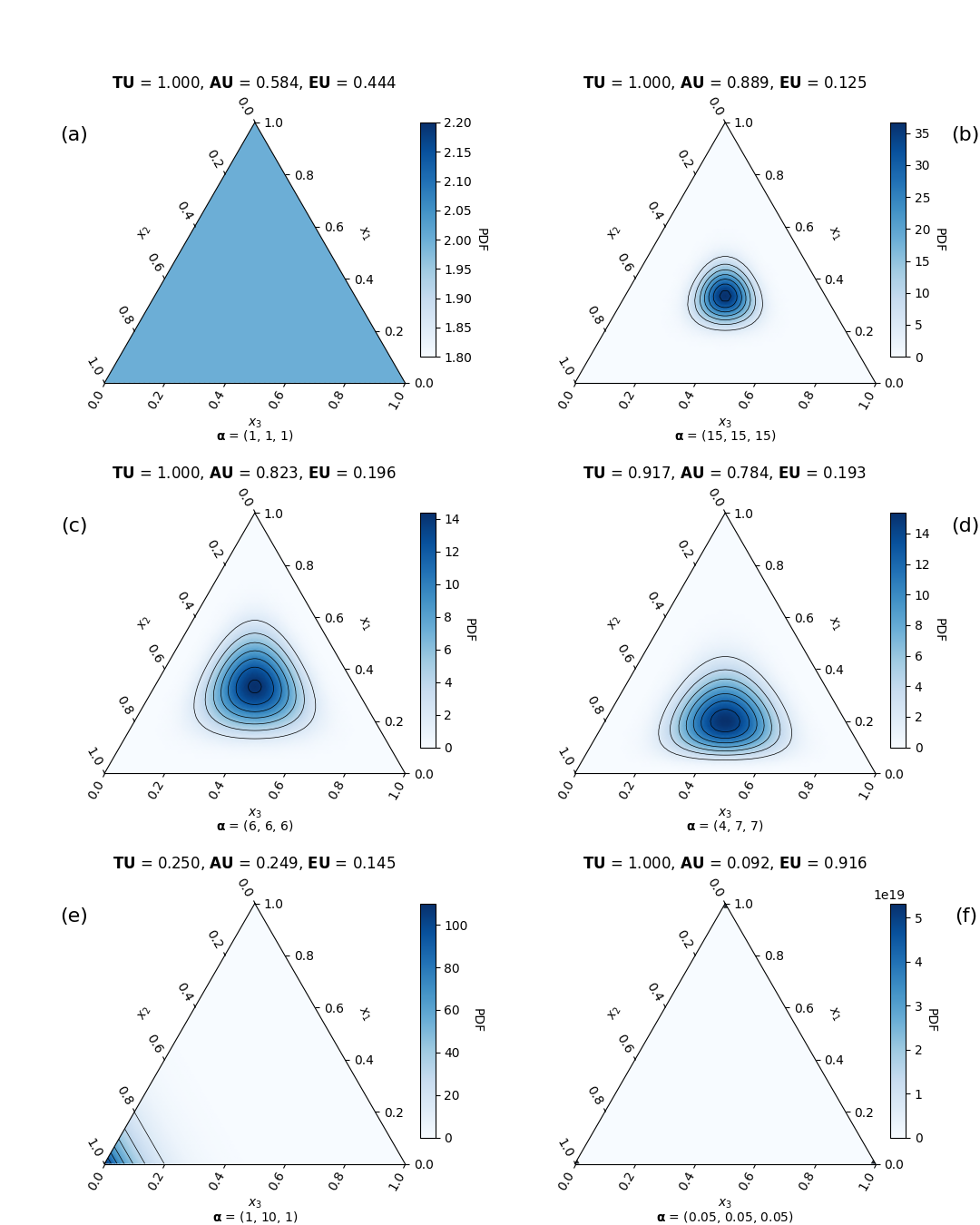}
		\caption{Dirichlet distributions with different choices of $\bm{\alpha}$ with normalized values for $\TU, \AU$, and $\EU$.}
		\label{fig2}
		\centering
	\end{figure}

	Figure \ref{fig2} displays, for $|\lab| = 3$, some exemplary Dirichlet distributions with different $\bm{\alpha}$ parameters over a 2-simplex, along with their corresponding normalized\footnote{The values are normalized by multiplying them with $K/(K-1),$ see discussion after Proposition \ref{prop:ranges}.} values for $\TU$, $\AU$, and $\EU$ in (\ref{tu:new}-\ref{eu:new}).
	We observe that the desired properties are captured as follows:
 	First, $\AU$ and respectively $\EU$ is always smaller or equal to $\TU$. Moreover, $\TU$ is maximal for the uniform distribution, as shown in Fig.\ \ref{fig2}a. 
    $\TU$ also attains its maximum under other parameter conditions, but with varying aleatoric and epistemic contributions:
    This can occur with a high $\AU$ value, stemming from a high concentration around the first-order uniform distribution (see Fig.\ \ref{fig2}b, c). Alternatively, a high $\EU$ value can drive this, due to a strong similarity to the discrete uniform distribution on the first-order Dirac measures, namely the vertices (see Fig.\ \ref{fig2}f).
    Additionally, we observe that $\EU$ strictly increases for mean-preserving spreads (Fig.\ \ref{fig2}b, c). 
    Fig.\ \ref{fig2}e depicts a Dirichlet distribution which is quite confident about one of the actual outcomes. This is reflected accordingly in low values of the uncertainty measures.
    These observations based on Dirichlet distributions align with our theoretical analysis of the proposed distance-based uncertainty measures.

	\section{CONCLUSION}
	\label{sec:conc}
		Recent criticisms have pointed to limitations in widely accepted uncertainty measures for second-order distributions, primarily due to certain unfavorable theoretical properties. Responding to this criticism, we presented a set of formal criteria that any uncertainty measure should fulfill. 
		Additionally, we introduced a distance-based approach to obtain uncertainty measures for total, aleatoric and epistemic uncertainties obeying the criteria.
		On the basis of the Wasserstein metric, we have demonstrated that this approach is fruitful and practical, especially for the often-used Dirichlet distributions.

	Our results open several venues for future work. 
	First, it would be interesting to instantiate the proposed uncertainty measures with metrics on probability measures other than the Wasserstein metric and verify that the proposed criteria are met. 
	In that respect, it would be interesting to work out general properties that a metric must satisfy in order for the criteria to be met.     
	Although the focus of our work is on the theoretical aspects of uncertainty measures, a systematic experimental comparison in the context of evidential deep learning would be intriguing. 

\subsubsection*{Acknowledgements}
Yusuf Sale is supported by the DAAD program Konrad Zuse Schools of Excellence in Artificial Intelligence, sponsored by the Federal Ministry of Education and Research. Michele Caprio would like to acknowledge partial funding by the Army Research Office (ARO MURI W911NF2010080). 

	\bibliography{references.bib}
	
	\clearpage

\onecolumn 

\appendix
\section{Proofs}

\subsection*{Proof of Lemma \ref{lemma_wassers}}
Since $\mathcal{Y}=\{y_1,\ldots,y_K\}$ is finite, this means that a probability measure $p \in \mathbb{P}(\mathcal{Y})$ can be seen as a $K$-dimensional probability vector. In symbols, $p \simeq (p(y_1),\ldots,p(y_K))^\top$. The latter is a vector belonging to the $(K-1)$-unit simplex $\Delta_{K-1}$. As a consequence, a second-order distribution on $\mathbb{P}(\mathcal{Y})$ can be seen as a first-order distribution on the $(K-1)$-unit simplex $\Delta_{K-1}$. In symbols, $\mathbb{P}(\mathbb{P}(\mathcal{Y})) \simeq \mathbb{P}(\Delta_{K-1})$. This, together with the first-order Wasserstein distance being a well-defined metric on $\mathbb{P}(\mathcal{Y})$ \citep{villani2009optimal,villani2021topics}, allows us to conclude that the second-order Wasserstein distance is itself a well-defined metric. \qed 

\subsection*{Proof of Proposition \ref{prop_optimal_coupling}}
Let $\delta_p \in \prob(\prob(\lab))$ be a second-order Dirac measure, where $p \in \prob(\lab)$, and $Q \in \prob(\prob(\lab))$ a second-order distribution. 
We show that any coupling $\gamma \in (\plab \times \plab, \sigma(\plab \times \plab)$ has to be necessarily given by $\gamma = Q \otimes \delta_p$. \\[0.2cm]
Thus, for any $A, B \in \sigma(\plab)$ we show that 
\begin{align*}
    \gamma(A,B) =
  \begin{cases}
    Q(A), & \text{if } p \in B, \\
    0, & \text{else}.
  \end{cases}
\end{align*}
Let $p \in B$, then we have $\gamma(A \times B^c) = 0$. Hence, this implies $\gamma(A \times B) = \gamma(A \times \plab) - \gamma(A \times B^c) = Q(A)$. This shows the first case. Assume $p \notin B$, then $\gamma(A \times B) \leq \gamma(\plab \times  B ) = \delta_p(B) = 0$, showing the second case. \qed 

\subsection*{Proof of Proposition \ref{prop:min}}
Let $p,q \in \plab$. Note that $ \gamma(y,y) = \min\{p(y), q(y)\}$ is trivially a coupling, hence $\gamma \in \Gamma(p, q)$. For the corresponding marginals we have $p(y) = \sum_{y^{\prime} \in \lab} \gamma(y,y^{\prime})$ and $q(y) = \sum_{y^{\prime} \in \lab} \gamma(y^{\prime},y)$, thus $\gamma(y, y) \leq \min\{p(y), q(y)\}$. This implies directly that $\min\{p(y), q(y)\}$ maximizes $\sum_{y \in \lab} \gamma(y,y)$, and therefore minimizes the distance $d_1(p,q) = \inf_{\gamma\in \Gamma(p, q)} \left\{ 1- \sum_{y \in \lab} \gamma(y,y) \right\}$. \qed 

\subsection*{Proof of Proposition \ref{prop:explicit_repr}}
Let the distance on $\plab$ be given by $d_1(p,q) = \frac{1}{2} \| p - q \|_1$, where $p,q \in \plab$. Now, for any $Q \in \pplab$ the proposed uncertainty measures simplify as follows:
\begin{align}
        \TU(Q) = \min_{y \in \lab} \mathcal{W}(Q, \delta_{\delta_y})  
           &= \min_{y \in \lab}  \mathbb{E}_{p \sim Q} \left[\frac{1}{2} \lVert p - \delta_y \rVert_1 \right]  \\[0.1cm]
           &=\min_{y \in \lab}  \mathbb{E}_{p \sim Q} \left[ 1 - p(y) \right] \label{dtv:dirac} \\[0.1cm]
           &= 1-\max_{y \in \lab} \mathbb{E}_{p \sim Q_y}[p(y)] .
\end{align}
Note that for \eqref{dtv:dirac} we used the fact that $\frac{1}{2} \lVert p - \delta_y \rVert_1 =\frac{1}{2} (1- p(y) + \sum_{y^{\prime} \neq y} p(y^{\prime}))= 1 - p(y)$.

Further, we have 
\begin{align}
    \AU(Q) &= \min_{\delta_m \in \Delta_{\delta_m} }\mathcal{W}(Q, \delta_m) \\[0.2cm]
     &= \min_{\delta_m \in \Delta_{\delta_m} } \inf_{\gamma \in \Gamma(Q, \delta_m)  } \int_{\prob(\lab) \times \prob(\lab)} d_1(p,q) \, d\gamma (p,q)  \\[0.2cm]
     &=\min_{\delta_m \in \Delta_{\delta_m} } \inf_{\gamma \in \Gamma(Q, \delta_m)  } \int_{\prob(\lab) \times \Delta_{\delta_y}} 1-p(y) \, d\gamma (p,q) \\[0.2cm]
     &\geq \int_{\prob(\lab)} 1-\max_{y \in \lab} p(y) \, dQ(p) \label{lower:bound} \\[0.2cm]
     &= \mathbb{E}_{p \sim Q}[1-\max_yp(y)],
\end{align}
where we used $y = \argmax_{y^{\prime} \in \lab} q(y^{\prime})$. Equality in \eqref{lower:bound} is reached for the Dirac mixture with $\delta_m^* \in \Delta_{\delta_m}$ with $\delta_m^*(\delta_y) = Q(y = \argmax_{y^{\prime} \in \lab} p(y^{\prime}))$. Thus, we have the following:
\begin{align*}
    \AU(Q) &=\inf_{\gamma \in \Gamma(Q, \delta_m^*)  } \int_{\prob(\lab) \times \Delta_{\delta_y }} 1-\sum_{y \in \lab} p(y)q(y)\,\,d\gamma (p,q) \\[0.2cm] 
    &= \inf_{\gamma \in \Gamma(Q, \delta_m^*)  } \int_{\prob(\lab)}\sum_{q \in \Delta_{\delta_y }} \{1-\sum_{y \in \lab} p(y)q(y)\}\gamma (q|p)\,\,dQ(p) \\[0.2cm] 
    &=\inf_{\gamma \in \Gamma(Q, \delta_m^*)  } \int_{\prob(\lab)}\sum_{y \in \lab} \{1-p(y)\}\gamma (\delta_y|p)\,\,dQ(p) \\[0.2cm] 
    &= \int_{\prob(\lab)} 1-\max_{y \in \lab} p(y) \, dQ(p) .
\end{align*}

The conditional probability measure $\gamma (\delta_y|p)  = \mathbbm{1}_{\{y =\argmax_{y^{\prime} \in \lab} p(y^{\prime}) \}}$ is valid, since
\begin{align*}
    \delta_m^*(\delta_y) =\int_{\prob(\lab)}\gamma (\delta_y|p)\, dQ(p) = Q(y = \argmax_{y^{\prime} \in \lab} p(y^{\prime})).
\end{align*}

Finally, we also have the following:
\begin{align*}
\EU(Q) &= \min_{\delta_p \in \Delta_{\delta_p} } \mathcal{W}(Q, \delta_p)  \\[0.1cm]
 &= \min_{\delta_p \in \Delta_{\delta_p}} \int_{\plab} \int_{\plab} d_1(p,q) \, d\delta_p(p) \, dQ(q) \\[0.1cm]
 &= \frac{1}{2} \min_{p \in \plab} \int_{\plab} \|q - p \|_1 \, dQ(q) \\[0.1cm]
 &= \frac{1}{2} \min_{p \in \plab} \mathbb{E}_{q \sim Q}[\| q - p \|_1] .
\end{align*}
This concludes the proof. \qed

\subsection*{Proof of Proposition \ref{prop:ranges}}
Let $Q \in \pplab$, then we have:
\begin{itemize}
    \item[i.)] $\TU(Q)= 1- \max_{y \in \lab} \mathbb{E}_{Q}[p(y)]   \leq 1-\frac{1}{K} = \frac{K-1}{K}$, where the inequality is a direct consequence of  $\sum_{y \in \lab} p(y) =1$ for any $p \in \mathbb{P}(\lab)$ which implies that $\max_{y \in \lab} p(y) \geq 1/K.$
    It is clear that this upper bound is reached for $Q^{\prime} \in \pplab$ such that $\mathbb{E}_{Q^{\prime}}[p] = \mathrm{Unif}(\lab)$.
    \item[ii.)] $\AU(Q) = 1 - \mathbb{E}_{Q}[\max_{y \in \lab} p(y)] \leq 1-\frac{1}{K} = \frac{K-1}{K}$. Clearly the upper bound is reached for $Q^{\prime} \in \pplab$ such that $Q^{\prime} = \delta_{\mathrm{Unif}(\lab)}$. 
    \item[iii.)] For $\EU(Q)$ we obtain 
\begin{align}
    \EU(Q) &= \frac{1}{2} \min_{p \in \plab} \mathbb{E}_{q \sim Q}[\| q - p \|_1]  \\[0.1cm]
    &\leq \frac{1}{2} \, \mathbb{E}_{q \sim Q}[\| q - \mathbb{E}_Q[p] \|_1] \\[0.1cm]
    &\leq \frac{1}{2} \, \mathbb{E}_{q \sim \delta_m}[\| q - \mathbb{E}_Q[p] \|_1] \label{eq:diracm} \\[0.1cm]
    &= \sum_{y \in \lab} \mathbb{E}_{Q}[p(y)] (1-\mathbb{E}_{Q}[p(y)] ) \\[0.1cm]
    &= 1- \sum_{y \in \lab} \mathbb{E}_{Q}[p(y)]^2 \\[0.1cm]
    &\leq 1-\frac{1}{K} = \frac{K-1}{K}, \label{eq:cauchy}
\end{align}
where \eqref{eq:diracm} follows from the Dirac mixture $\delta_m \in \Delta_{\delta_m}$ being a mean-preserving spread of $Q$. Inequality \eqref{eq:cauchy} is a consequence of the Cauchy-Schwarz inequality, and the linearity of expectation. The upper bound is reached for $Q^{\prime},$ which is such that $Q^{\prime}(\delta_y) = \frac{1}{K}$ for all $y \in \lab$. 
\end{itemize}
This concludes the proof. \qed

\subsection*{Proof of Corollary \ref{corallary:31}}
Corollary \ref{corallary:31} is an immediate consequence of Proposition \ref{prop:ranges}. \qed

\subsection*{Proof of Theorem \ref{theorem:axioms}}
We show that the proposed uncertainty measures  with the Wasserstein distance instantiation satisfy A0-A9 defined in Section 2.2.
\begin{itemize}
    \item[A0:] Since the proposed measures are distance based this property holds trivially true.
    \item[A1:] Let $p \in \plab$ and $y \in \lab$, then we have
\begin{align*}
\AU(\delta_{\mathrm{Unif}(\lab)}) &= 1-\max_{y \in \lab} \mathrm{Unif}(\lab)(y) \\
&= \frac{K-1}{K} \\
&\geq 1-\max_{y \in \lab} p(y) \\
&= \AU(\delta_p) \\
&\geq 0 \\
&= 1 - \max_{y' \in \lab} \delta_y(y') \\
&= \AU(\delta_{\delta_y}).
\end{align*}
The first inequality is a direct consequence of Proposition \ref{prop:ranges}.
    \item[A2:] For $p \in \plab$ and $Q \in \pplab$ we have immediately by definition $\EU(Q) \geq \EU(\delta_p) = 0$. The other inequality in this axiom follows directly from Proposition \ref{prop:ranges} iii.). 
    \item[A3:] Since $\bigcup_{y \in \lab}\{\delta_{\delta_y}\} \subset \Delta_{\delta_p}$ it follows $\EU(Q) =  \min_{\delta_p \in \Delta_{\delta_p}} \mathcal{W}(Q,\delta_p) \leq  \min_{y \in  \lab}  \mathcal{W}(Q,\delta_{\delta_y}) = \TU(Q)$, for any $Q \in \pplab$.
    Similarly, since $\bigcup_{y \in \lab}\{\delta_{\delta_y}\} \subset \Delta_{\delta_m}$ we obtain $\AU(Q) =  \min_{\delta_p \in \Delta_{\delta_m}} \mathcal{W}(Q,\delta_m) \leq  \min_{y \in  \lab}  \mathcal{W}(Q,\delta_{\delta_y}) = \TU(Q)$, for any $Q \in \pplab$.
    \item[A4:] This follows from Proposition \ref{prop:ranges} (i), since we have $\mathbb{E}_{Q}[p] = \mathrm{Unif}(\lab)$ for $Q$ being the continuous second-order uniform distribution.
    \item[A5:]  Further, let $Q^{\prime} \in \pplab$ be a mean-preserving spread of $Q \in \pplab$, i.e., let $X \sim Q, X^\prime \sim Q^\prime$ be two random variables such that $X^\prime \overset{d}{=} X + Z$, for some random variable $Z$ with $\mathbb{E}[Z| X = x] = 0$, for all $x$ in the support of $X$.     
    Then, we have 
    \begin{align*}
    \EU(Q') &= \frac{1}{2} \min_{p \in \plab} \mathbb{E}[\lVert (X + Z) - p \rVert_1] \\[0.1cm]
           &= \frac{1}{2} \min_{p \in \plab}\sum_{i = 1}^{K} \mathbb{E}(| X_i + Z_i - p_i|) ,
\end{align*}
    where $X_1,\ldots,X_K$ are the marginals of $X$ and $Z_1,\ldots,Z_K$ the marginals of $Z$, respectively.
    From this, we further infer that for any $p = (p_1, \dots, p_k) \in \plab$
 and any $x$ in the support of $X$ that
 \begin{align*}
          \EU(Q') 
           &\geq \frac{1}{2} \min_{p \in \plab}\sum_{i = 1}^{K} \mathbb{E}(| X_i - p_i| + Z_i (\mathbbm{1}_{X_i>p_i } - \mathbbm{1}_{X_i<p_i}) \\[0.1cm]
           &=\frac{1}{2} \min_{p \in \plab}\sum_{i = 1}^{K} \mathbb{E}_{Q_i}(\mathbb{E}_{Z_i}(| X_i - p_i| + Z_i (\mathbbm{1}_{X_i>p_i } - \mathbbm{1}_{X_i<p_i}) \, | \, X_i = x_i)) \\[0.1cm]
           &=\frac{1}{2} \min_{p \in \plab}\sum_{i = 1}^{K} \mathbb{E}_{Q_i}(| X_i - p_i| + \mathbb{E}(Z_i \, | \,X_i = x_i)(\mathbbm{1}_{X_i>p_i } - \mathbbm{1}_{X_i<p_i})) \\[0.1cm]
           &=\frac{1}{2} \min_{p \in \plab} \sum_{i = 1}^{K} \mathbb{E}_{Q_i}(| X_i - p_i|)  \\
           &= \EU(Q).
\end{align*}
    \item[A6:] Let $Q$ be a second-order distribution with mean $p \in \plab$. Assume that $Q^{\prime} \in \pplab$ is a spread-preserving shift of $Q$, shifted along the vector $z$ with $\sum_{i = 1}^{K} z_i = 0$. Then, we obtain 
\begin{align*}
    \EU(Q^{\prime}) &= \frac{1}{2} \min_{(p + z) \in \plab} \mathbb{E}_{q \sim Q^{\prime}}[\| q - (p + z) \|_1] \\[0.1cm]
    &= \frac{1}{2} \min_{(p + z) \in \plab} \mathbb{E}_{q \sim Q}[\| (q + z) - (p + z) \|_1] \\[0.1cm]
    &= \frac{1}{2} \min_{p \in \plab} \mathbb{E}_{q \sim Q}[\| q  - p  \|_1] \\[0.1cm]
    &= \EU(Q).
\end{align*}
    \item[A7:] Let $\lab_1$ and $\lab_2$ be partitions of $\lab$ and $Q \in \prob(\prob(\lab))$. Further, denote by $Q_{|\lab_i}$ the marginalized distribution on $\lab_i$. First, we observe that $\mathbb{E}_{Q}[p] = (\mathbb{E}_{Q|\lab_1}[p], \mathbb{E}_{Q|\lab_2}[p])^\top$. This observations yields 
    \begin{align*}
        \TU(Q) &=  1-\max_{y \in \lab} \mathbb{E}_{p \sim Q_y}[p(y)] \\[0.1cm]
        &=  1- \max\left(  \max_{y_1 \in \lab_1} \mathbb{E}_{Q|\lab_1}[p(y_1)]   \, , \, \max_{y_2 \in \lab_2} \mathbb{E}_{Q|\lab_2}[p(y_2)] \right). 
    \end{align*}
    From this, we immediately see that 
    $$ \TU(Q) \leq 1-  \max_{y_1 \in \lab_1} \mathbb{E}_{Q|\lab_1}[p(y_1)] = \TU_{\lab_1}(Q_{|\lab_1}) $$
    as well as 
    $$  \TU(Q) \leq 1-  \max_{y_2 \in \lab_2} \mathbb{E}_{Q|\lab_2}[p(y_2)] =\TU_{\lab_2}(Q_{|\lab_2}) . $$
    This implies $  \TU(Q) \leq \TU_{\lab_1}(Q_{|\lab_1})  + \TU_{\lab_2}(Q_{|\lab_2})$ as asserted.
    \item[A8:] This property follows immediately, since $\TU$ only depends on the mean of the respective second order distribution $Q \in \pplab$.
\end{itemize}
This concludes the proof. \qed

\subsection*{Proof of Proposition \ref{prop:eu_dirichlet}}
Assume that $Q \in \pplab$ is given by a Dirichlet distribution $\Dir(\bm{\alpha})$, so that the marginal distributions are Beta distributions, i.e., $Q_i \sim \text{Beta}(\alpha_i, \alpha_0 - \alpha_i)$ with $\alpha_0 = \sum_{j=1}^{K} \alpha_j$ for each $i \in \lab$. Hence, we seek to solve the following constraint optimization problem:
\begin{align}
		&\underset{q \in (0,1)^K}{\textbf{minimize}} \quad h(q) \coloneqq \frac{1}{2} \sum\nolimits_{i=1}^{K} \mathbb{E}_{p_i \sim Q_i} [\, | p_i - q_i | \,] \label{opt:obj} \\[0.3cm]
		&\textbf{subject to} \quad c(q) \coloneqq \sum\nolimits_{i = 1}^{K} q_i - 1 = 0
		\label{opt:const}
\end{align}

Further evaluation of the terms in the objective function yields:
\begin{align*}
     \mathbb{E}_{Q_i}(| p_i - q_i |) &= \int_{0}^{1}|p_i - q_i|f(p_i)\,dp_i \\
                          &= \int_{q_i}^{1}(p_i - q_i)f(p_i)\,dp_i + \int_{0}^{q_i}(q_i-p_i)f(p_i)\,dp_i \\
                          &= \int_{q_i}^{1}p_i f(p_i)\,dp_i - \int_{0}^{q_i}p_i f(p_i)\,dp_i + 
                         q_i\int_{0}^{q_i}f(p_i)\,dp_i - q_i\int_{p_i}^{1}f(p_i)\,dp_i \\
                         &= \int_{q_i}^{1}p_i f(p_i)\,dp_i - \int_{0}^{q_i}p_i f(p_i)\,dp_i + q_iF(q_i) - q_i(1-F(q_i))\\
                         &= \int_{q_i}^{1}p_i f(p_i)\,dp_i - \int_{0}^{q_i} p_i f(p_i)\,dp_i + q_i(2F(q_i) - 1).
\end{align*}
It is easy to evaluate the involved integrals, since we know
\begin{align*}
     \int_{a}^{b}p_if(p_i)\,dp_i &= \frac{1}{B(\alpha, \beta)} \int_{a}^{b} p_i^\alpha(1-p_i)^{\beta-1} \,dp_i\\  
     &= \frac{B(\alpha+1,\beta)}{B(\alpha, \beta)} \int_{a}^{b} \frac{1}{B(\alpha+1,\beta)} p_i^{(\alpha+1)-1} (1-p_i)^{\beta-1}\,dp_i \\
     &= \frac{\alpha}{\alpha + \beta} P_{\alpha+1, \beta}(a \leq p_i \leq b).
\end{align*}

Together, this yields 
\begin{align*}
    \EU(Q) &= \frac{1}{2} \mathbb{E}_Q(\lVert p - q \rVert_1)\\ &= \frac{1}{2}\sum_{i = 1}^{K} \mathbb{E}_{Q_i}(| p_i - q_i |) \\
    &= \frac{1}{2}\sum_{i = 1}^{K} \left[\frac{\alpha_i}{\alpha_0}(1-2F_{\alpha_i+1, \alpha_0}(q_i)) + q_i(2F_{\alpha_i,\alpha_0}(q_i) - 1) \right] .
\end{align*}

Further, the \textit{Lagrangian function} is given by
	\begin{align}
		\mathcal{L}(q, \lambda) &= h(q) + \lambda \, c(q),
		\label{eq:lag1}
	\end{align}
	where $\lambda \in \mathbb{R}$. The extreme points of the Lagrangian \eqref{eq:lag1} are the solutions of the equations 
	\begin{align}
		\nabla h(q) = -\lambda \nabla c(q) \label{lag:det} \\[0.2cm] 
		\sum\nolimits_{i = 1}^{K} q_i - 1 = 0.
	\end{align}

The solution to \eqref{lag:det} is given by
\begin{align*}
 F_{\alpha_i,\alpha_0}(q_i) - \frac{1}{2} =  -\lambda \qquad \forall i=1,\ldots,K \\[0.2cm] 
 \Leftrightarrow q_i = F^{-1}_{\alpha_i,\alpha_0}(\frac{1}{2} - \lambda) \qquad \forall i=1,\ldots,K.
\end{align*}
Since the cdf is strictly monotone increasing there can be no more than one solution under the constraint \eqref{opt:const}. The quantile function is continuous and for $\lambda \rightarrow 0.5$ it becomes $0$ and for $\lambda \rightarrow -0.5$ it goes towards $1$, hence there must be exactly one $\lambda \in \mathbb{R}$ such that both equations \eqref{opt:obj} and \eqref{opt:const} are fulfilled. Clearly, the bordered Hessian is given by: 
\begin{align}
    H(q,\lambda) = \begin{bmatrix} 
    0 & 1 & \dots &1\\
    1 & f_{\alpha_1,\alpha_0}(q_1) & \dots & 0 \\
    \vdots & \vdots & \ddots & \vdots\\
    1 & 0&   \dots    & f_{\alpha_K,\alpha_0}(q_K)
    \end{bmatrix}
\end{align}

Consequently, for the determinant of the bordered Hessian we get
\begin{align*}
   \det(H(q,\lambda)) &= -\frac{\prod_if_{\alpha_i, \alpha_0}(q_i)}{\sum_if_{\alpha_i, \alpha_0}(q_i)}.
\end{align*}
Since there is only one solution and the determinant of the bordered Hessian is negative, we conclude that the minimum is unique. Thus, the constraint optimization problem has a unique solution. \qed

\end{document}